\documentclass[12pt]{article}

\usepackage[margin=1in]{geometry}
\usepackage{setspace}
\usepackage{tabularx}
\usepackage{array}
\usepackage{float}
\usepackage{graphicx}
\usepackage{url}

\title{Hidden Clues in WM-811K:\\
Uncovering the Spatial Signals Behind Wafer Map Prediction}

\author{Kabir Tolani}
\date{}

\begin{document}

\maketitle
\begin{abstract}

WM-811K (also known as LSWMD) is a wafer map dataset first made publicly available in a 2015 paper by Wu et al.

Convolutional neural networks have been used to achieve high classification accuracy on this dataset, yet it's not easy to decipher which features of the dataset are driving that performance.

This study attempts to answer this specifically for WM-811K: how much of the predictive performance results from the wafer layout, the count of failed dies, their distance from the center and their exact 2-D location.

After analyzing duplicate wafer maps, removing same wafer maps with conflicting labels and ensuring that the same manufacturing lot or exact same map doesn’t appear in training, validation and test data, we were left with 25,513 defect-labeled wafer maps from eight classes.

We then constructed four variations of each wafer map. The Original was the actual wafer map present in the dataset. The Layout-only kept only the layout of the non-background dies. The Global-shuffle kept the count of failed dies while shuffling them around. The Radial-shuffle randomized the locations of failed dies only in their radial bin, which was chosen out of 8 radial bins based on distance from the center.

The data split, preprocessing and CNN architecture remained the same for all four of them.

Macro-F1 progressed like this: 0.3829 (Layout-only), 0.5803 (Global-shuffle), 0.7439 (Radial-shuffle) and 0.8563 (Original).

These results show that WM-811K contains useful information at several levels: wafer layout, number of failures, distribution of failed dies in a given radius, and exact 2-D locations of the failed dies. Therefore, high classification performance on this benchmark does not come only from learning the exact defect configuration.

\end{abstract}

\section{Introduction}

A typical semiconductor wafer is part of a lot. A lot has multiple wafers. A wafer has
multiple dies (chips). Wafers are circular, whereas the dies placed on them are
rectangular. A production lot is supposed to have 25 wafers, and in our copy, the median
lot size was 24. The wafer maps are stored as rectangular 2D arrays and the additional
pixels required for this padding are zeroes (background) \cite{wu2015,kagglewm811k}.

Researchers over the years have used multiple machine learning approaches to study the
WM-811K dataset. CNNs are especially popular since they work very well with visual data.
A literature survey shows that a lot of that research has been on higher classification
accuracy and more efficient neural networks \cite{kim2023review,zhou2026review,nakazawa2018,shin2023}. Some papers have focussed on handcrafted
features and quantum neural networks as well \cite{kang2021,prabhu2026,hu2026,marwaha2026}.

However, CNNs are not easily interpretable, which makes it difficult to figure out whether
they learned the correct features intended by the researcher \cite{selvaraju2017}. CNNs might accidentally use
a hidden correlation, which is known as shortcut learning \cite{geirhos2020}. Or some data might be present
across the train/test boundary, which is known as data leakage \cite{kaufman2012}.

For wafers in the WM-811K dataset, there might be easy clues present for the CNNs to
learn. For example, a particular lot might be prone to certain defects. Many of the wafer
maps share the exact same layout.

A study showed that visually persuasive heatmaps may not accurately describe the
model’s actual decision process \cite{jia2026}. Another study on heatmap-based explanations found that
sometimes the region highlighted in the Grad-CAM heatmap was not where the actual
defect lay \cite{selvaraju2017,dudeja2026}.

In this study, we do not create new handcrafted features. Nor do we train a very
complicated SOTA classifier. Rather, we remove and then add spatial information back in
a graded manner. The main question we are trying to answer is how much incremental
performance can be achieved by the count of failed chips, their approximate distance from
the center, and the exact 2D arrangement on top of the wafer layout.

To the best of our knowledge, there is no WM-811K study that ensures that production lots
and exact wafer maps are distinct across training, validation, and test splits and at the
same time measures the performance across 4 different spatial arrangements as
explained above \cite{kim2023review,zhou2026review,dudeja2026,kim2024polar,hu2026}.

\section{Methods}

\subsection{Dataset and initial audit}

The study used the LSWMD.pkl version of WM-811K. It had 811,457 rows, each of which
had 6 columns. One of those columns was a wafer map, which was a 2D array and each
of its values was 0,1 or 2. 0 meant background (no die), 1 meant a healthy die, and 2
meant a failed die. Another column was dieSize, and ideally, the dieSize for a wafer-map
should be equal to count of 1s and 2s in the wafer map. This was experimentally found to
be true for all wafer maps. FailureType denoted the type of error present, and only 8 of
those - Center, Donut, Edge-Loc, Edge-Ring, Loc, Near-full, Random, and Scratch - were
of interest to the study. The rest were either unlabeled or labeled ‘none’. Wafer maps
labeled ‘none’ might contain failed dies but are considered free of a systematic error
pattern. Furthermore, unlabeled wafers are those that weren’t reviewed by human experts.
The WaferIndex column denoted the position of the wafer in its corresponding lot \cite{wu2015,kagglewm811k}.

The other two columns were lotName and trianTestLabel (misspelt). Using these two, we
ascertained that the lot names were distinct in the supplied test/train split, i.e., none of the
lots were present in both the test/train rows. However, we found exact wafer maps present
in both the test and train sets, which impedes our ability to use the supplied test/train split
as it would give an unfair advantage to the model. This led us to create Strict Groups,
which are explained in the next subsection. Also, all duplicate wafer map clusters had the
same labels except 3. In those 3 clusters, each of size 2, both the wafer maps had
different labels. We decided to drop them from our dataset rather than choosing one label
at random.
\subsection{Strict groups and leakage-aware splitting}

The strict group as a concept was introduced to tackle the problem of the same wafer map
being present across train/test boundaries. Wafers from the same lot belong to the same
strict group. If any two lots share the exact same wafer map, all wafers from those two lots
will belong to the same strict group. So, it’s like building a chain of connected lots and
putting them in the same strict group. If lot 1 and lot 2 share a wafer map, and lot 2 and lot
3 share a wafer map, all three lots will move to the same strict group. This method resulted
in the 25,513 wafers being split across 7,952 strict groups.

Any time data was split into train/test/validation roles, wafers from each strict group stayed
together in one of those roles only. Model weights were learned using a training set. The
best training epoch was computed using a validation set. Final evaluation was conducted
using the test set.

Before model training, we compared three-fold and five-fold grouped cross-validation.
Three folds were selected because they gave better support for the rare classes,
especially Near-full. The frozen train/validation/test sizes were 13,794/3,375/8,344 in fold
1, 13,509/3,465/8,539 in fold 2, and 13,598/3,285/8,630 in fold 3. No test data were used
to choose the validation split.
\clearpage
\subsection{Four representations of the wafer map}
Our experiment uses four representations of every wafer map. Table~\ref{tab:four-representations} shows a quick
summary.

\begin{table}[H]
\centering
\caption{The four representations and the question they help answer.}
\label{tab:four-representations}
\begin{tabularx}{\textwidth}{
    >{\raggedright\arraybackslash}p{0.19\textwidth}
    >{\raggedright\arraybackslash}X
    >{\raggedright\arraybackslash}X
    >{\raggedright\arraybackslash}X
}
\hline
Representation &
Information kept &
Information destroyed &
Main question \\
\hline

\textbf{Layout-only} &
Layout of the valid dies &
Failure count and failure location &
What is the influence of wafer layout on failure label prediction? \\

\hline

\textbf{Global-shuffle} &
Layout + failure count &
Failure location &
What is the incremental performance gain with failure count? \\

\hline

\textbf{Radial-shuffle} &
Layout + total failure count + failure count in each radial band &
Exact location of a failure in its 2D radial band &
What is the incremental gain with an approximate circular map of failures? \\

\hline

\textbf{Original} &
All information as present in the original wafer map &
No destruction &
What is the incremental gain with exact failure location? \\

\hline
\end{tabularx}
\end{table}

\begin{figure}[H]
\centering
\includegraphics[width=\textwidth]{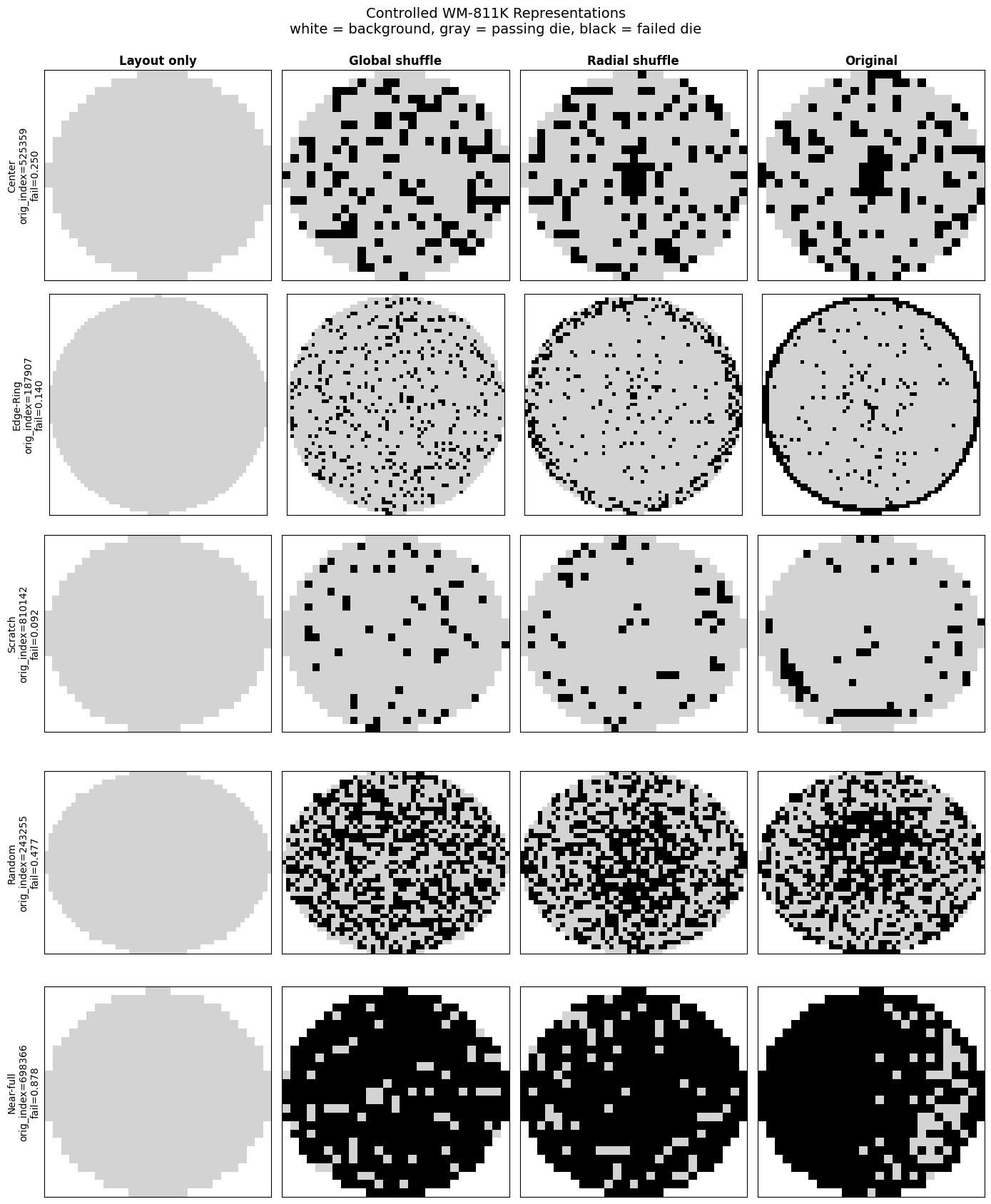}
\caption{Examples of Layout-only, Global-shuffle, Radial-shuffle, and Original representations.}
\label{fig:four-representations}
\end{figure}

To convert a wafer map to Layout-only, we changed all 2s to 1s while leaving the 0s
unchanged. So, the background pixels didn’t change; only the valid die pixels all became
the same. For producing the Global-shuffle version, again, we didn’t touch the background
pixels; we just distributed failures across the valid die layout while keeping their count the
same. Similarly, for Radial-shuffle, the failure count in a given radial band stayed the same,
with their positions shuffled in the same radial band. We used eight equal-width radial
bands for this. Figure~\ref{fig:four-representations} shows examples of the four representations.

To ensure that each wafer map always produces the same randomized version, each
wafer map got its custom random seed derived from the master seed 42, the wafer map’s
exact hash, and other parameters. This ensured that the order of processing of wafer
maps didn’t matter, and the same wafer maps would always receive the same randomized
version.

\subsection{Image preprocessing}

Each wafer image was converted to a 160x160 image. We audited several image sizes,
and another strong candidate was 224x224, which was large enough to accommodate
each of the 25,513 images without downscaling. But choosing that size would result in
nearly twice the pixels (50,176 vs. 25,600), and we were working in a constrained
computing environment on Google Colab. Also, 160x160 was able to fit 99.91\% (25,489 of
25,513) of the images without downscaling, so it was a practical trade-off.

Those 24 downscaled samples were noted, and later, the evaluation was repeated after
excluding them.

For downscaling the images, we preserved the aspect ratio and used nearest-neighbor
interpolation. For the smaller images, we put them in the center and padded them with
zeroes to make them 160x160. They were not enlarged.

In the original wafer maps, each pixel was 0/1/2, as noted earlier. But feeding them as is to
a CNN carried the risk that they might be interpreted as numeric quantities, and “2” might
be seen as twice “1”. To prevent that, each pixel was replaced with two binary channels.
The first value denoted 0 for background and 1 for valid die. The second one was 0 for
healthy and 1 for failing. So, the three valid values were [0,0] for background, [1,0] for
healthy die, and [1,1] for failed die.

\subsection{CNN architecture and training}

We used the same CNN architecture for each of the four representations, because the
primary thing we wanted to assess was changes in predictive performance as incremental
information was added. Also the architecture was reasonably simple and conventional as
we were not chasing SOTA performance. Our CNN had 4 convolutional blocks which had
16, 32, 64, and 128 features, respectively.

Each block had 2 3x3 convolutional layers, so effectively we had 8 convolutional layers.
Each block also had max pooling preceded by ReLU activation and batch normalization.

We chose Flattening over GAP (Global average pooling) so that we can preserve more
spatial information, which would be likely to be destroyed if we used GAP. For example,
we wanted to differentiate between whether the high defect density is at the center or
around the periphery.

Once flattened, the 10 x 10 x 128 vector became a 12,800 x 1 vector. Then it passed
through a 256-unit dense layer, 30\% dropout and 8-class softmax output. The CNN had
around 3.5 million trainable parameters. Supplementary Figure~\ref{fig:supp-cnn-architecture} shows the complete
architecture.

We chose AdamW as the optimizer \cite{loshchilov2019} as it’s a common reliable choice for CNNs. It keeps
the model weights small and prevents overfitting. We used a learning rate of $3\times 10^{-4}$ and
weight decay of $1\times 10^{-4}$.

The model looked at 64 wafers at a time and was allowed to go through the training
dataset a maximum of 40 times to train itself (batch size 64, training length: 40 epochs). If
the model did not improve its performance for 8 epochs, we stopped the training, known as
patience of 8 epochs. We kept saving the weights whenever the model gave the best
macro-F1 (checkpointing).

We did not rotate/flip our images to create more data for training (no data augmentation).

We used balanced class weights so that the CNN is penalized more for its mistakes in rare
classes, which helps it to fix its mistakes faster. If the error penalty is not weighted like this,
the CNN might “cheat” by giving more importance to large classes.

We used 3 different model initialization seeds (42, 123 and 456). Why? When a CNN model’s
training starts, the model chooses some random numbers as its weights. Choosing a
different seed results in different initial weights. Doing it 3 times reduces the chance that
we simply got lucky with a certain set of random numbers.

Hence, our primary experiment ran 36 times, as we ran it once for each combination of 4
representations x 3 folds x 3 seeds.

\subsection{Evaluation and statistical analysis}

We chose macro-F1 as the main metric. F1 combines recall and precision. Macro-F1 is a
simple average of per class F1 values. Therefore, a class with fewer samples gets the
same weightage as a larger class. We also captured ordinary and balanced accuracy
among others.

Since we split our data using 3-Fold StratifiedGroupKFold \cite{pedregosa2011}, we later combined predictions
from all the 3 test sets in such a way that each wafer was counted only when it was part of
the test fold and never when it was part of the training data.

In order to compute the 95\% confidence interval on the results, we computed macro-F1
2,000 times by doing resampling with replacement. We already had prediction results
made by our CNNs for each wafer. For example, if the strict groups in a test fold were A,
B, C, D, and E, we might get A, B, B, E, and D, or A, A, C, C, and D, if we do resampling
with replacement. This is called bootstrap resampling \cite{efron1979}.

\section{Results}

\subsection{Dataset audit}

During the dataset audit, we determined that no lots were present in both test and train
subsets. Therefore, the test/train split was lot-disjoint. However, it was not exact-wafer-
map-disjoint since 92 exact maps were present in both. Also 3 duplicate pairs had
conflicting labels (Supplementary Figure~\ref{fig:supp-conflicting-maps}). After removing them, we were left with 25,513, down from the original
25,519.

This audit was useful since exact wafer maps in both the test and train subsets might
result in inflated performance \cite{kaufman2012}. Our usage of strict groups removed this problem and
allowed us to focus on the primary experiment of incremental spatial information addition.

\subsection{Each addition of spatial information resulted in clear performance gains}

\begin{figure}[H]
\centering
\includegraphics[width=\textwidth]{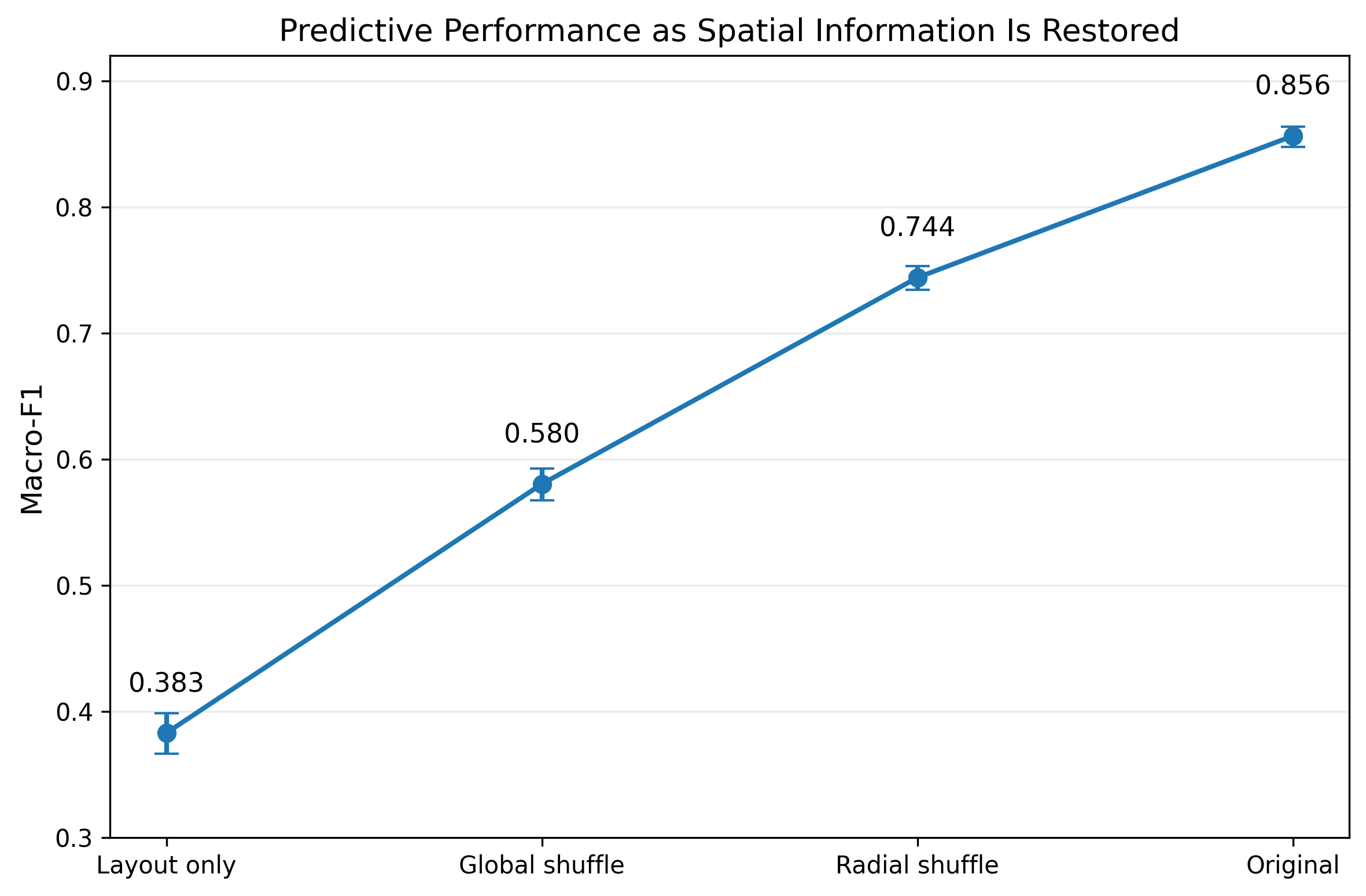}
\caption{Macro-F1 as spatial information is incrementally restored with 95\% Confidence interval using
bootstrap resampling.}
\label{fig:macro-f1-progression}
\end{figure}

Performance progression was clearly visible at every step of incremental failure
information addition. Macro-F1 moved as 0.3829, 0.5803, 0.7439, and 0.8563 as the
representation changed from Layout-only, Global-shuffle, Radial-shuffle and Original. Figure~\ref{fig:macro-f1-progression}
shows the 95\% confidence intervals derived using strict group bootstrap resampling.

Table~\ref{tab:representation-performance} summarizes the performance of all four wafer map
representations across the reported evaluation metrics.

\begin{table}[H]
\centering
\caption{Performance of all four wafer map representations}
\label{tab:representation-performance}
\begin{tabularx}{\textwidth}{
    >{\raggedright\arraybackslash}p{0.22\textwidth}
    >{\centering\arraybackslash}p{0.15\textwidth}
    >{\centering\arraybackslash}p{0.15\textwidth}
    >{\centering\arraybackslash}p{0.15\textwidth}
    >{\centering\arraybackslash}p{0.16\textwidth}
    >{\centering\arraybackslash}p{0.13\textwidth}
}
\hline
Condition & Macro-F1 & 95\% CI low & 95\% CI high & Balanced accuracy & Accuracy \\
\hline
Layout-only & 0.3829 & 0.3664 & 0.3987 & 0.4713 & 0.5518 \\
Global-shuffle & 0.5803 & 0.5672 & 0.5925 & 0.6101 & 0.6363 \\
Radial-shuffle & 0.7439 & 0.7344 & 0.7528 & 0.7639 & 0.7925 \\
Original & 0.8563 & 0.8475 & 0.8636 & 0.8665 & 0.8953 \\
\hline
\end{tabularx}
\end{table}

Other metrics, such as Accuracy and Balanced Accuracy showed similar trends.
Layout-only had a non-trivial performance on its own, while the Original had the best
numbers.

\subsection{Incremental spatial information gains}

Since we designed our experiment as a spatial information ladder, the incremental gain
in the macro-F1 at each step can be interpreted as the predictive value of the information
added at each step. Therefore, the predictive values of failed die count, approximate radial
location, and exact 2D location are 0.1974, 0.1636, 0.1123.

\begin{figure}[H]
\centering
\includegraphics[width=\textwidth]{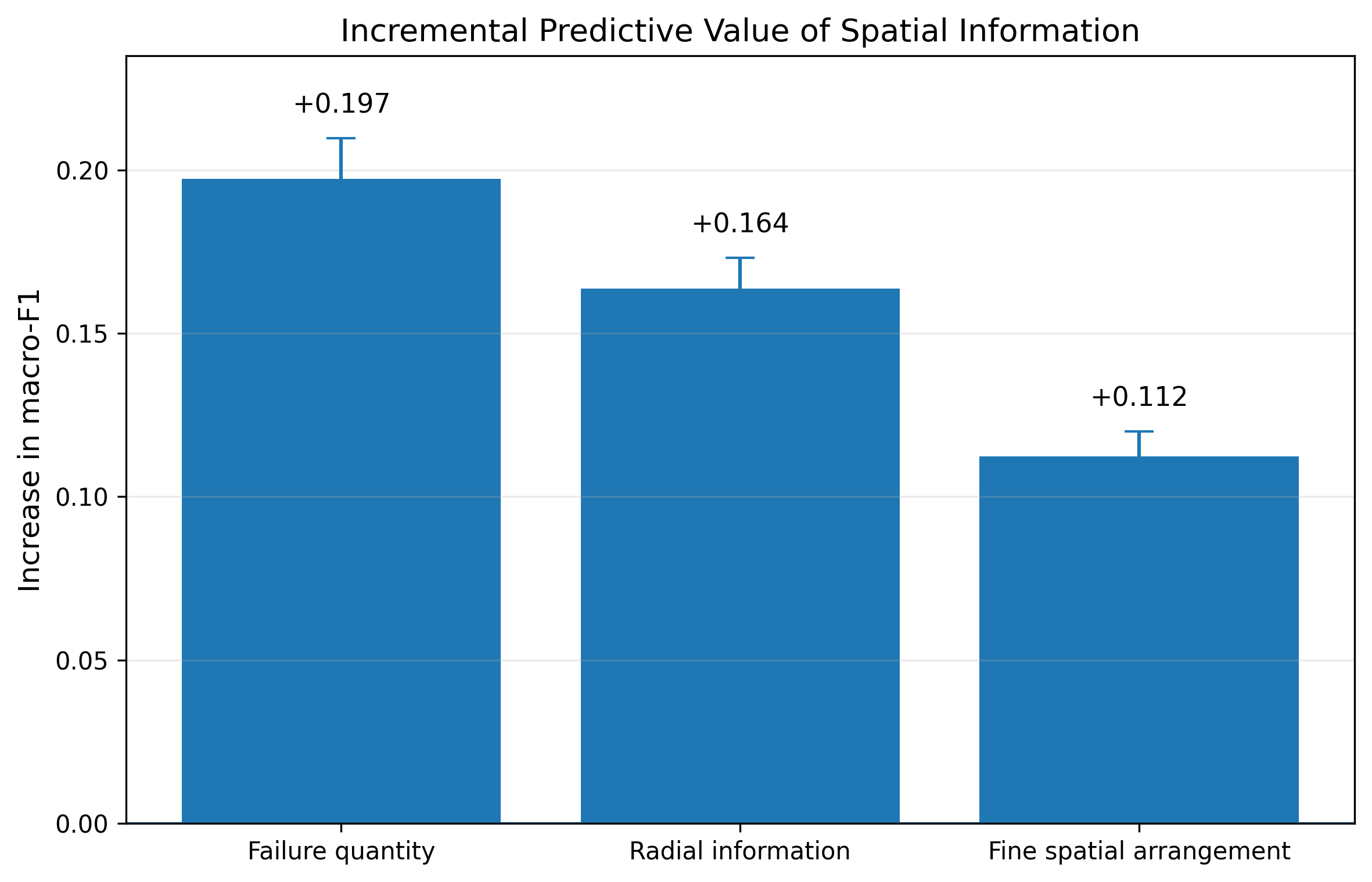}
\caption{Additional macro-F1 increments from each of the 3 steps. Error bars denote 95\% confidence intervals.}
\label{fig:incremental-gains}
\end{figure}

Figure~\ref{fig:incremental-gains} visualizes these incremental gains, while
Table~\ref{tab:incremental-gains} reports the corresponding values and 95\% confidence intervals.

\begin{table}[H]
\centering
\caption{Incremental macro-F1 gains and 95\% confidence intervals.
Gains were calculated using unrounded values; displayed values are rounded.}
\label{tab:incremental-gains}
\begin{tabularx}{\textwidth}{
    >{\raggedright\arraybackslash}p{0.30\textwidth}
    >{\centering\arraybackslash}p{0.18\textwidth}
    >{\centering\arraybackslash}p{0.18\textwidth}
    >{\centering\arraybackslash}p{0.18\textwidth}
}
\hline
Information added & Macro-F1 gain & 95\% CI low & 95\% CI high \\
\hline
Failure quantity & +0.1974 & +0.1860 & +0.2097 \\
Radial information & +0.1636 & +0.1551 & +0.1731 \\
Fine spatial
arrangement & +0.1123 & +0.1049 & +0.1199 \\
\hline
\end{tabularx}
\end{table}

All 3 confidence intervals above zero mean that even if we take the lower end of
performance gains, they are positive. Other metrics, including Balanced and ordinary
accuracy, showed similar trends.

Balanced-accuracy gains were +0.1389, +0.1538, and +0.1026, while ordinary-accuracy gains were
+0.0845, +0.1562, and +0.1028.
\subsection{Different defect labels utilized different spatial information}

Defect label (class) specific results show which spatial information is useful for which
class. Near-full is a good example. The jump to .948 (Global-shuffle) from .051 (Layout-
only) shows that failure count was very useful in predicting whether a defect wafer
was Near-full. Furthermore, later information additions weren’t meaningful, detrimental
even.

The Center and Donut classes benefited heavily from the radial failure count information.
Their scores showed large jumps from Global-shuffle to Radial-shuffle. They are
naturally related to failure distance from the center.

For Scratch, we see a large jump from Radial-shuffle to Original, showing that
exact failure defect location is important to identify a pattern which is made up of a
continuous directional line.

\begin{figure}[H]
\centering
\includegraphics[width=\textwidth]{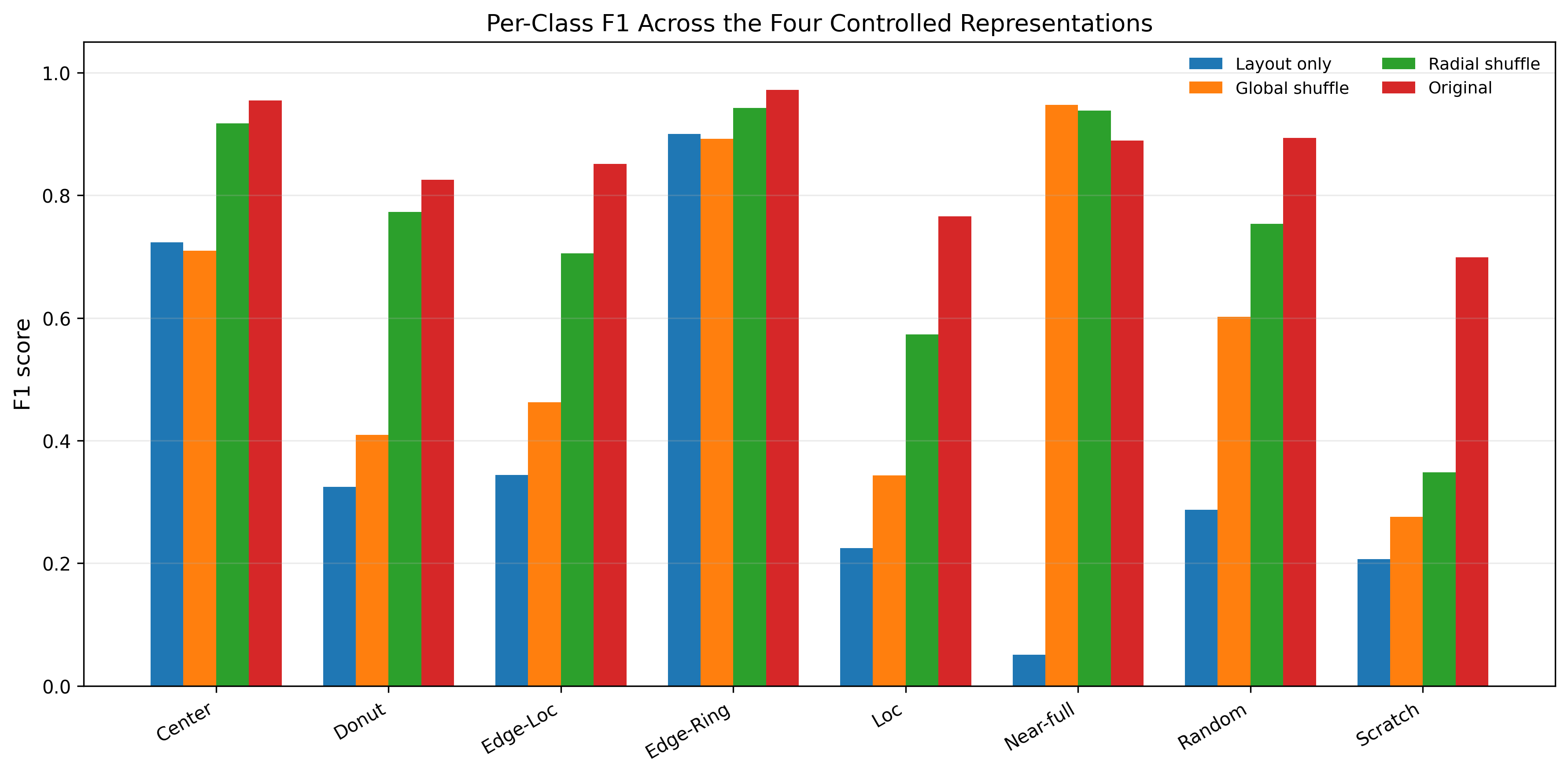}
\caption{Per-class F1 for the four representations.}
\label{fig:per-class-f1}
\end{figure}

Figure~\ref{fig:per-class-f1} visualizes the per-class F1 scores across the four
representations, while Table~\ref{tab:per-class-f1} provides the corresponding numerical values.

\begin{table}[H]
\centering
\caption{Per-class F1 for the four representations.}
\label{tab:per-class-f1}
\begin{tabularx}{\textwidth}{
    >{\raggedright\arraybackslash}p{0.20\textwidth}
    >{\centering\arraybackslash}p{0.18\textwidth}
    >{\centering\arraybackslash}p{0.18\textwidth}
    >{\centering\arraybackslash}p{0.18\textwidth}
    >{\centering\arraybackslash}p{0.18\textwidth}
}
\hline
Defect class & Layout-only & Global-shuffle & Radial-shuffle & Original \\
\hline
Center & 0.723 & 0.710 & 0.917 & 0.955 \\
Donut & 0.325 & 0.409 & 0.773 & 0.825 \\
Edge-Loc & 0.344 & 0.462 & 0.706 & 0.851 \\
Edge-Ring & 0.900 & 0.892 & 0.943 & 0.972 \\
Loc & 0.225 & 0.343 & 0.573 & 0.766 \\
Near-full & 0.051 & 0.948 & 0.938 & 0.889 \\
Random & 0.287 & 0.602 & 0.754 & 0.893 \\
Scratch & 0.207 & 0.276 & 0.348 & 0.699 \\
\hline
\end{tabularx}
\end{table}

Not every class shows incremental gains with every richer representation. Near-full is a
good example where performance gains are actually in reverse order starting from the
second step. However, the broader trends are stable.

\subsection{Performance stability under different random seeds and without downscaled samples}

The performance ranking for the four layouts was the same for all three model seeds (Supplementary Figure~\ref{fig:supp-seed-consistency}). Macro-F1 values were 0.3829 $\pm$ 0.0028 for Layout-only, 0.5803 $\pm$ 0.0044 for Global-shuffle,
0.7439 $\pm$ 0.0043 for Radial-shuffle, and 0.8563 $\pm$ 0.0099 for Original maps. Here the $\pm$ values represent the standard deviation across the three seeds.

The exclusion of the 24 wafer samples that had to be downscaled due to our choice of a
160x160 resolution did not meaningfully affect the numbers.

\section{Discussion}

\subsection{Layout, failure count, approximate radial distance and exact 2D location - all were
important}

One surprising result was a macro-F1 of 0.3829 from the wafer layout alone. A
classifier making a random guess based on frequency of each class would get a macro-F1
value of .125.

For classes such as Edge-Ring, where intuitively the radial distance should have had a
significant predictive performance, the intuition turned out to be correct. Similarly, the exact
2D location mattered for Scratch.

Also worth mentioning is that, for Near-full, the predictive performance of failure count plus layout 
was more than that of the case when radial distribution and then  exact defect location were restored.

This is possibly because shuffling evened out the noise present, and the classifier found it
easy to predict Near-full on uniformly distributed error map.

\subsection{Potential evidence of dataset shortcut in performance of Layout-only}

We saw that macro-F1 of .3829 was achieved simply using the valid die layout.
Therefore, there was a strong correlation between the wafer layout and the defect class. A
possible explanation could be that the manufacturing process of certain wafer layouts is
prone to certain defects.

Also, this is possibly a dataset shortcut. For example, if we wanted to classify cat and dog
images, but accidentally cat pictures always had a blue background while dog pictures
always had a green background. Consequently, the model learns a shortcut to associate
green backgrounds with dogs without actually learning what a dog looks like \cite{geirhos2020}.

Also worth noting is why we allowed the same layouts to be present in the test and train
datasets. We actually wanted to measure how much predictive performance is due to the
layout alone. Therefore, we did not keep the test/train split layout-disjoint as we did when
making it lot-disjoint and exact-wafer-map-disjoint.
\subsection{Number of failed chips and their approximate distance from center explain most of the performance gain}

Adding the count of failed dies atop the wafer layout resulted in the maximum jump of
+0.1974 in the macro-F1 score. The explanation is simplest for Near-full. If most of the
dies are already failed, the layout and the number of failures are enough to identify the
class. Also, during the dataset audit and later shuffle analysis, we came across 14 Near-full
wafers which were unchanged by Global-shuffle and Radial-shuffle. This is because all their
dies were failed (Supplementary Figure~\ref{fig:supp-near-full-unchanged}).

A large gain of +0.1636 was obtained by restoring the coarse radial distribution. This is
highly expected for the classes where failures tend to occur at specific locations. Examples
include the Center, Donut, Edge-Loc, and Edge-Ring classes. Recent work also suggests that
radial information helps in wafer classification \cite{kim2024polar,prabhu2026,hu2026}.
\subsection{Exact defect location is important}

The experiment shows that the exact pixel by pixel defect location is actually quite
important. Even after restoring the layout, failure count, and approximate radial distance,
restoring the exact 2D location was still able to add +0.1123 in terms of the macro-F1
score, which is significant.

The evidence is strongest for the Scratch class. Its score increased from 0.348 to 0.699.
The most likely explanation is that Scratch is mostly defined as a continuous line segment
which gets destroyed in the Radial-shuffle.
\subsection{Implications for WM-811K benchmarking}

Many wafer map studies focus on achieving high classification accuracy \cite{kim2023review,zhou2026review,shin2023}. However, in
the absence of controlled information addition experiments, such as the one in this study, it
would be hard to say with confidence whether the corresponding model used any
shortcuts.

Hence, a step-by-step information addition experiment would be a useful complement
to standard classification accuracy. One can speculate that a model which
emerges better in those graded tasks, will perform better on the unseen datasets, possibly
from a different factory.
\subsection{Limitations}
\begin{enumerate}
\item The study uses only one public dataset WM-811K. Therefore, the same results may
not reproduce on other datasets.

\item The four representations may not be truly independent. It is possible that the radial
location, failed die count, and wafer layout depend on each other. In that case, the
incremental predictive gains with each layer of information addition may not
represent the quantum of information contained in the dataset.

\item We used only one CNN architecture. A more robust study would have used different
CNNs and even different families of classifiers.

\item The classes are highly imbalanced. Consequently, the results for rare classes such
as Near-full, which contain only 149 wafers, are likely to be unstable.
\end{enumerate}
\section{Conclusion}

This project aimed to determine which set of information in the wafer map was
responsible for driving CNN prediction performance, while keeping the classifier the same
across the different representations of the same wafer map. The results were clear
in the sense that macro-F1 score kept increasing with each additional layer of information.
The achieved values were 0.3829 for Layout-only, 0.5803 for the failure count in addition to layout, and 0.7439 while using an approximately similar radial location in addition to layout and failure count. The Original maps achieved 0.8563, which was the highest score, as expected.

The study concludes that the exact 2D arrangement of the wafer defects is important,
but a non-trivial level of performance can be derived using wafer layout alone. Additionally,
adding the total failure count and their approximate radial arrangement yields further
meaningful performance improvements.

Controlled experiments such as this can reveal whether the underlying model used
certain correlations present in the dataset rather than the features which the experimenter
might have expected it to learn.

\section{Data and Reproducibility}

The primary dataset named, LSWMD.pkl, was downloaded from Kaggle \cite{kagglewm811k} and had a
SHA-256 checksum of
1d04fccb3dd3176b276878b926b20fead7e077c5751e4d353ea9741a5e7b5c65. After
cleaning, the dataset had 25,513 wafer maps distributed among 7,952 strict groups. The
central experiment was run 36 times (3 outer folds, 3 random seeds, 4 representations). A
total of 2,000 strict-group bootstrap resamples were used for statistical analysis. The Python
code for the study is available upon request.

\section*{Acknowledgments}

We would like to thank the reviewers who provided feedback on the project’s central
question and experiment methodology. All code was written in Python and run on Google
Colab. Multiple LLMs were used for coding, debugging, literature search, and planning.

\section*{Supplementary Figures}
\setcounter{figure}{0}
\renewcommand{\thefigure}{S\arabic{figure}}

\begin{figure}[H]
\centering
\includegraphics[width=\textwidth]{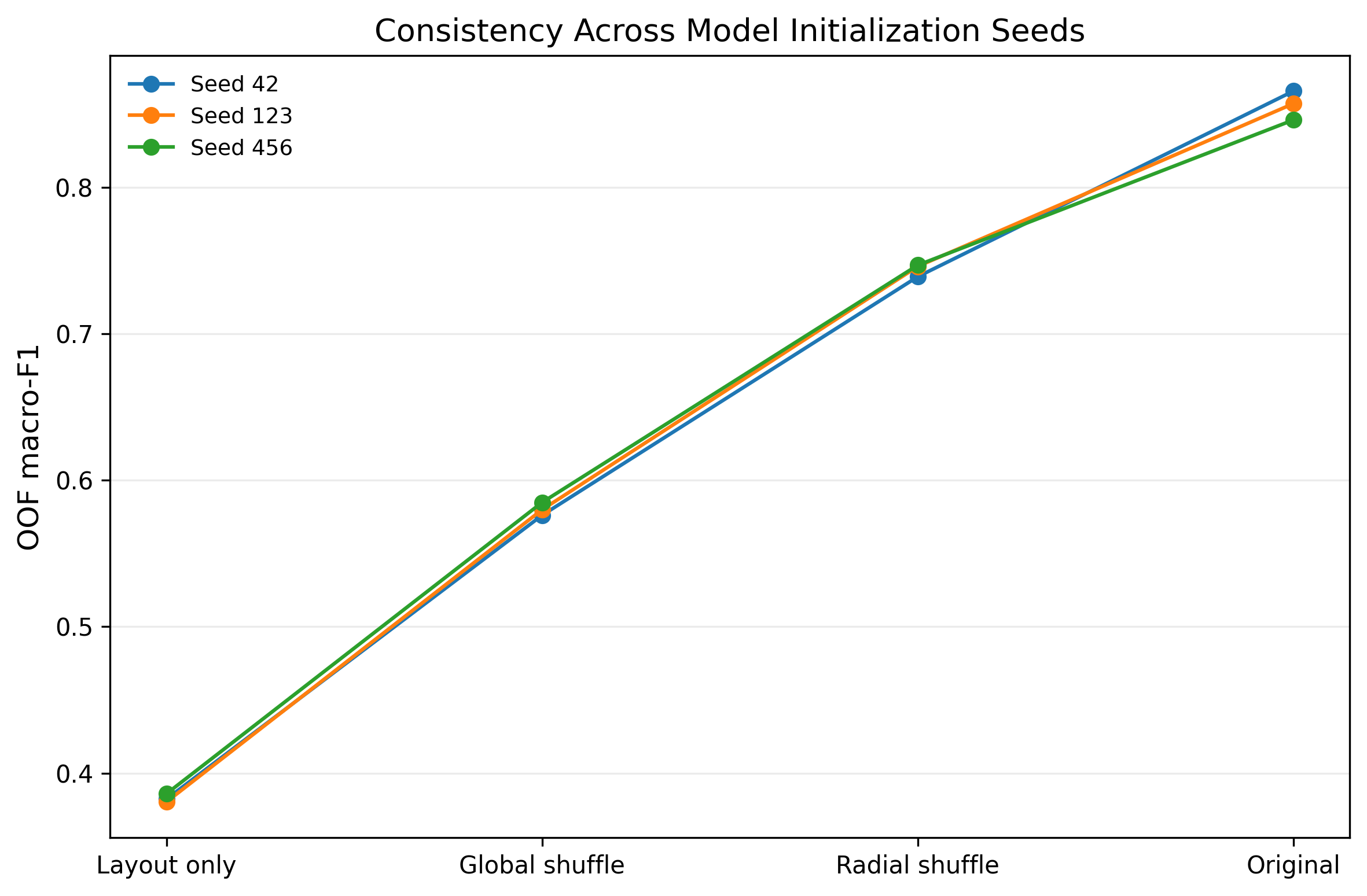}
\caption{Pooled out-of-fold macro-F1 for each model initialization seed. All three seeds show the same ordering:
Layout-only $<$ Global-shuffle $<$ Radial-shuffle $<$ Original.}
\label{fig:supp-seed-consistency}
\end{figure}
\begin{figure}[H]
\centering
\includegraphics[width=\textwidth]{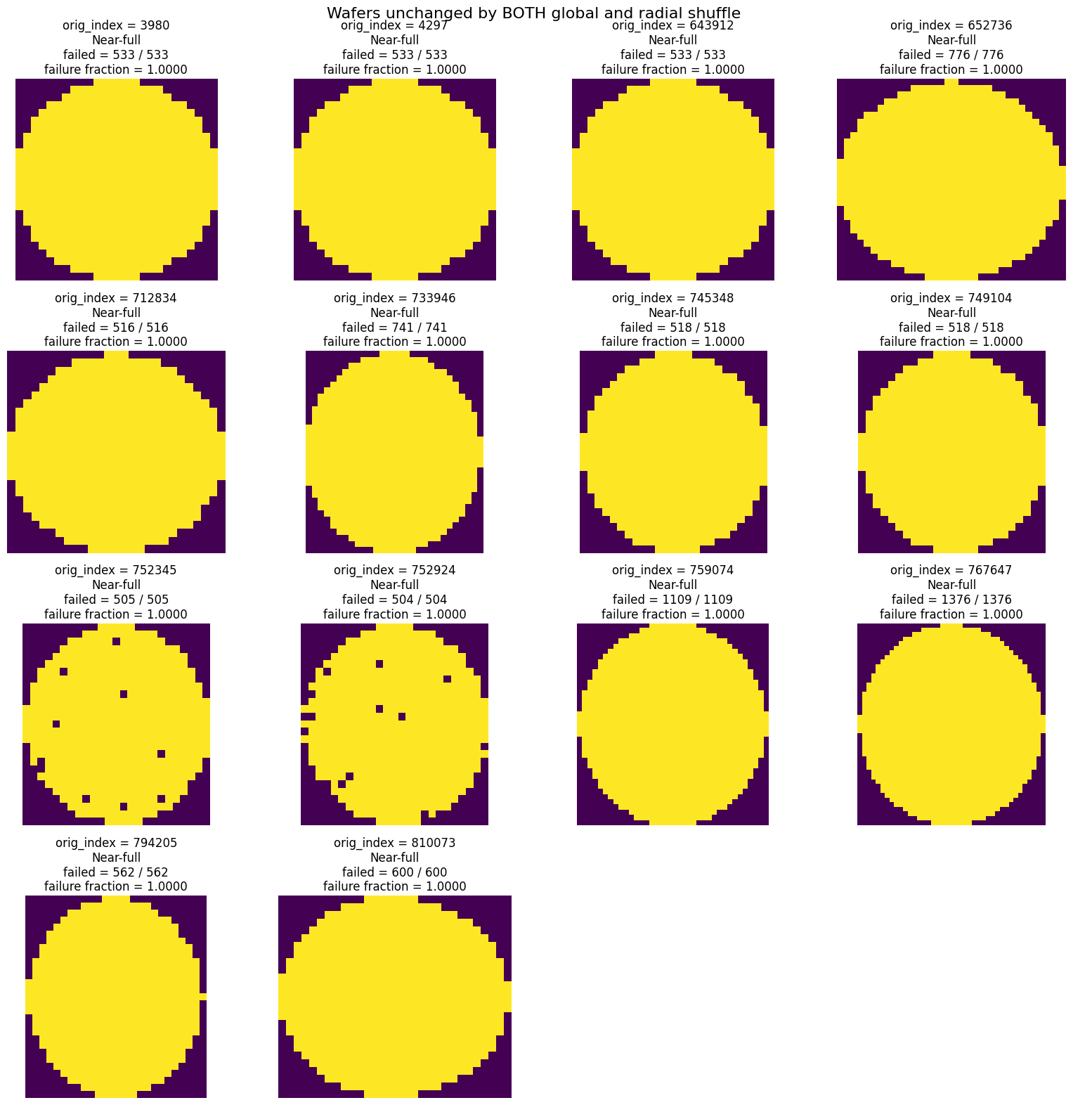}
\caption{Fourteen Near-full wafers were unchanged under both Global-shuffle and Radial-shuffle. These cases had failure
fraction = 1.000, meaning every valid die was already failed. Because the shuffles preserve the total number of failed
dies, and Radial-shuffle also preserves the failed-die count within each radial band, these maps could not be altered by
either randomization.}
\label{fig:supp-near-full-unchanged}
\end{figure}
\begin{figure}[H]
\centering
\includegraphics[width=\textwidth]{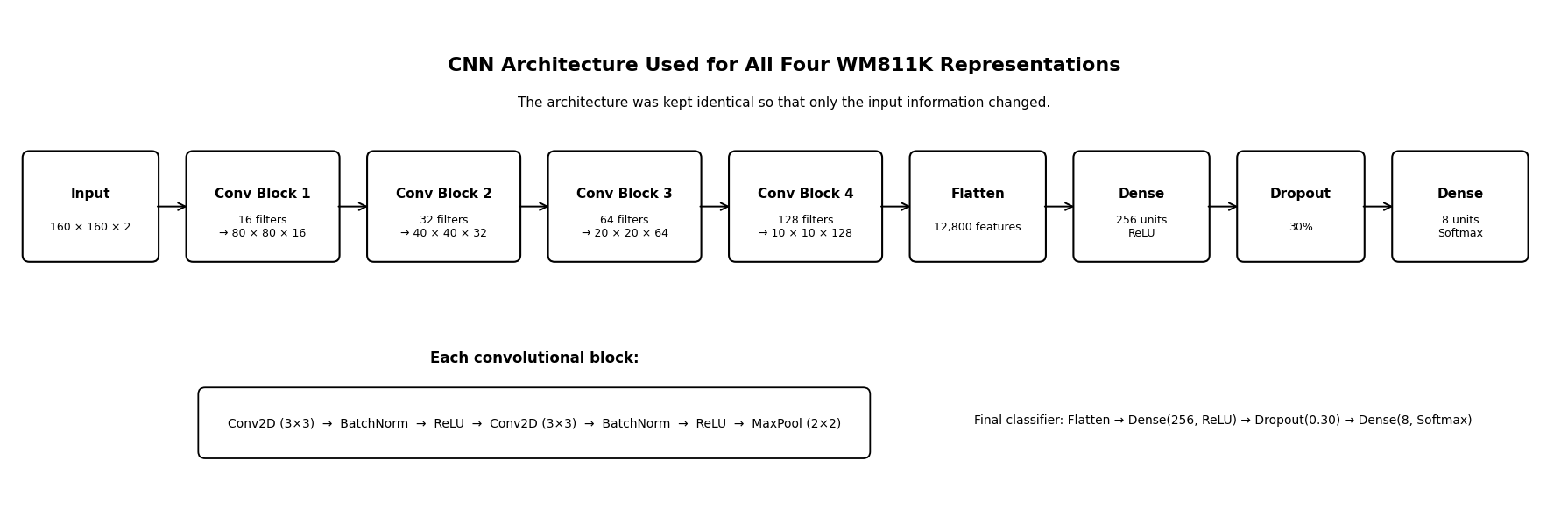}
\caption{CNN architecture used for all four controlled WM811K representations. Each convolutional block contains
two 3 $\times$ 3 convolutions, each followed by batch normalization and ReLU, and ends with 2 $\times$ 2 max pooling. The same
architecture was used for Layout-only, Global-shuffle, Radial-shuffle, and Original inputs.}
\label{fig:supp-cnn-architecture}
\end{figure}
\begin{figure}[H]
\centering
\includegraphics[width=\textwidth]{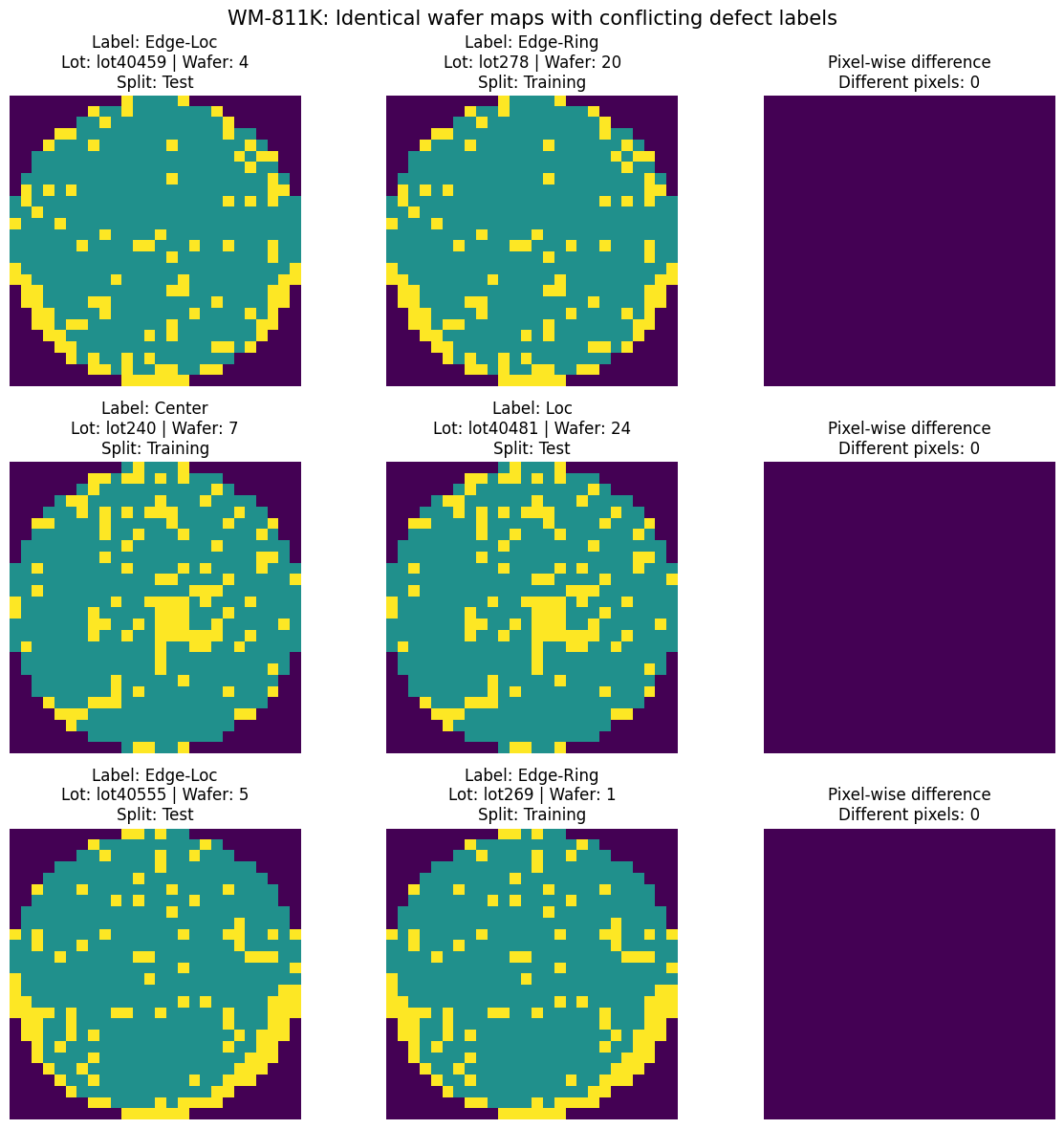}
\caption{Pixel-identical wafer maps with conflicting defect labels.}
\label{fig:supp-conflicting-maps}
\end{figure}

\end{document}